\documentclass[11pt,a4paper]{article}
\usepackage{authblk}
\usepackage[hyperref]{naaclhlt2018}
\usepackage{times}
\usepackage{graphicx}
\usepackage{latexsym}
\usepackage{array}
\usepackage{lipsum}
\usepackage [english]{babel}
\usepackage [autostyle, english = american]{csquotes}
\MakeOuterQuote{"}
\usepackage{url}
\usepackage{diagbox}
\usepackage{soul}

\aclfinalcopy 

\makeatletter
\newcommand\email[2][]%
   {\newaffiltrue\let\AB@blk@and\AB@pand
      \if\relax#1\relax\def\AB@note{\AB@thenote}\else\def\AB@note{\relax}%
        \setcounter{Maxaffil}{0}\fi
      \begingroup
        \let\protect\@unexpandable@protect
        \def\thanks{\protect\thanks}\def\footnote{\protect\footnote}%
        \@temptokena=\expandafter{\AB@authors}%
        {\def\\{\protect\\\protect\Affilfont}\xdef\AB@temp{#2}}%
         \xdef\AB@authors{\the\@temptokena\AB@las\AB@au@str
         \protect\\[\affilsep]\protect\Affilfont\AB@temp}%
         \gdef\AB@las{}\gdef\AB@au@str{}%
        {\def\\{, \ignorespaces}\xdef\AB@temp{#2}}%
        \@temptokena=\expandafter{\AB@affillist}%
        \xdef\AB@affillist{\the\@temptokena \AB@affilsep
          \AB@affilnote{}\protect\Affilfont\AB@temp}%
      \endgroup
       \let\AB@affilsep\AB@affilsepx
}
\makeatother

\title{Performance Comparison of Crowdworkers and NLP Tools on Named-Entity Recognition and Sentiment Analysis of Political Tweets}

\author[*]{Mona Jalal}
\author[**]{Kate K. Mays}
\author[**]{Lei Guo}
\author[*]{Margrit Betke}
\affil[*]{Department of Computer Science, Boston University}
\affil[**]{Division of Emerging Media Studies, Boston University} 
\email{\url{{jalal, kkmays, guolei, betke}@bu.edu}}

\begin{document}
\maketitle
\begin{abstract}

We report results of a comparison of the accuracy of crowdworkers and seven Natural Language Processing (NLP) toolkits in solving two important NLP tasks,  named-entity recognition (NER) and entity-level sentiment (ELS) analysis.  We here focus on a challenging dataset, 1,000 political tweets that were collected during the U.S. presidential primary election in February 2016.  Each tweet refers to at least one of four presidential candidates, i.e., four named entities.  The groundtruth, established by experts in political communication, has entity-level sentiment information for each candidate mentioned in the tweet.  We tested several commercial and open source tools.  Our experiments show that, for our dataset of political tweets, the most accurate NER system, Google Cloud NL, performed almost on par with crowdworkers, but the most accurate ELS analysis system, TensiStrength, did not match the accuracy of crowdworkers by a large margin of more than 30 percent points.
\end{abstract}

\section{Introduction}
As social media, specially Twitter, takes on an influential role in presidential elections in the U.S., natural language processing of political tweets 
\citep{MohammadZhKiMa15}
has the potential to help with nowcasting and forecasting of election results as well as identifying the main issues with a candidate -- tasks of much interest to journalists, political scientists, and campaign organizers \citep{FarzindarIn15}. As a methodology to obtain training data for a machine learning system that analyzes political tweets, 
\citet{SamekiGeMaGuBe16} devised a crowdsourcing scheme with variable crowdworker numbers based on the difficulty of the annotation task. They provided a dataset of tweets where the sentiments towards political candidates were labeled both by experts in political communication and by crowdworkers who were likely not domain experts. \citet{SamekiGeMaGuBe16} revealed that crowdworkers can match expert performance relatively accurately and in a budget-efficient manner.  Given this result, the authors envisioned future work in which groundtruth labels would be crowdsourced for a large number of tweets and then used to design an automated NLP tool for political tweet analysis. 

The question we address here is: How accurate are existing NLP tools for  political tweet analysis?  These tools would provide a baseline performance that any new machine learning system for political tweet analysis would compete against.  
We here explore  whether existing NLP systems can answer the questions \textbf{\textit{"What sentiment?"}} and \textbf{\textit{"Towards whom?"}} accurately for the dataset of political tweets provided by \citet{SamekiGeMaGuBe16}. In our analysis, we include NLP tools with publicly-available APIs, even if the tools were not specifically designed for short texts like tweets, and, in particular, political tweets.    

Our experiments reveal that the 
task of entity-level sentiment analysis is difficult for existing tools to answer accurately while the recognition of the entity, here, which politician, was easier.

\section{NLP Toolkits}

NLP toolkits typically have the following capabilities: tokenization, part-of-speech (PoS) tagging, chunking, named entity recognition and sentiment analysis. In a study by \citet{PintoOlAl16}, it is shown that the well-known NLP toolkits NLTK \citep{Bird06}, Stanford CoreNLP \citep{ManningSuBaFiBeMc14}, and TwitterNLP \citep{RitterClMaEt11} have tokenization, PoS tagging and NER modules in their pipelines.  

There are two main approaches for NER: (1) rule-based and (2) statistical or machine learning based. The most ubiquitous algorithms for sequence tagging use Hidden Markov Models~\citep{JurafskyMa08}, Maximum Entropy Markov Models~\citep{JurafskyMa08, McCallumFrPe00}, or Conditional Random Fields~\citep{SuttonMc12}. Recent works \citep{WangHuZhZh16, ZhangLi17} have used recurrent neural networks with attention modules for NER.

Sentiment detection tools like SentiStrength \citep{ThelwallBuPa10} and TensiStrength \citep{Thelwall17} are rule-based tools, relying on various dictionaries of emoticons,  slangs, idioms, and ironic phrases, and set of rules that can detect the sentiment of a sentence overall or a targeted sentiment. Given a list of keywords, TensiStrength (similar to SentiStrength) reports the sentiment towards selected entities in a sentence, based on five levels of relaxation and five levels of stress.

Among commercial NLP toolkits (e.g., \citet{aylien,msftnlp,watsonnlu}), we selected \citet{gcnlp} and \citet{rosette} for our experiments, which, to the best of our knowledge, are the only publicly accessible commercial APIs for the task of entity-level sentiment analysis that is agnostic to the text domain. We also report results of TensiStrength~\citep{Thelwall17}, TwitterNLP~\citep{RitterClMaEt11}, \citet{spaCy}, CogComp-NLP~\citep{KhashabiSaZhre18}, and Stanford NLP NER~\citep{FiGrMa05}.

\section{Dataset and Analysis Methodology}
\label{sec:dataset}

We used the 1,000-tweet dataset by \citet{SamekiGeMaGuBe16} that contains the named-entities labels and entity-level sentiments for each of the four 2016 presidential primary candidates Bernie Sanders, Donald Trump, Hillary Clinton, and Ted Cruz, provided by crowdworkers, and by experts in political communication, whose labels are considered groundtruth.  The crowdworkers were located in the US and hired on the \citet{mTurk} platform. For the task of entity-level sentiment analysis, a 3-scale rating  of "negative," "neutral," and "positive" was used by the annotators.


\begin{figure*}[ht]
\centering
\includegraphics[width=\textwidth]{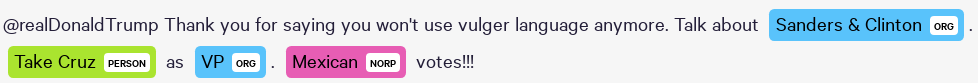}
\begin{center}
\line(1,0){450}
\end{center}
\vspace*{-0.2cm}
\includegraphics[width=0.9\textwidth]{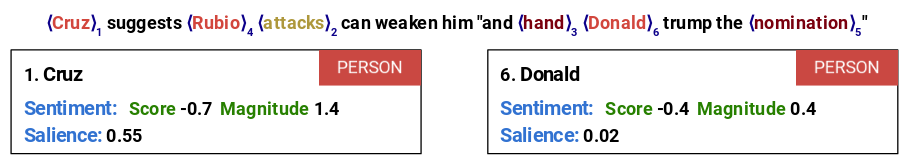}
\vspace*{-0.4cm}
\caption{Incorrect NER by spaCy (top) and incorrect ELS analysis by Google Cloud (bottom)}
\label{fig1}

\vspace{-.2cm}
\end{figure*}

\begin{table*}[ht]
\begin{center}
{\small 
\centering
\caption{Average Correct Classification Rate (CCR) for named-entity recognition (NER) of four presidential candidates and entity-level sentiment (ELS) analysis by NLP tools and crowdworkers}
\vspace{0.2cm}
\begin{tabular}{ll|c|cccc}
   & &  All Entities & Bernie Sanders & Donald Trump & Hillary Clinton & Ted Cruz \\
   \hline
 & \# Tweets & 1,000 & 236 & 510 & 225 & 211 \\
  \hline
NER & Rosette Text Analytics & \textbf{77.2\%} & 80.1\% & 79.8\% & 84.4\% & 60.2\% \\ 
& TwitterNLP & \textbf{81.2\%} & 89.4\% & 76.5\% & 79.6\% & 85.3\%  \\ 
& CogComp-NLP & \textbf{82.6\%} & 83.9\% & 81.2\% &  83.9\% & 79.6\%  \\

 & Stanford NER & \textbf{83.2\%} & 86.4\% & 76.8\% & 86.2\% & 83.4\%  \\
 & spaCy & \textbf{88.2\%} & 91.1\% & 72.7\% & 92\% & 91.5\%  \\
 & Google Cloud NL & \textbf{96.7\%} & 97.9\% & 96.1\% & 96.4\% & 97.1\% \\ 
 & mTurk crowdworkers & \textbf{98.6\%} & 100\% & 99\% & 97.3\% & 98.1\%  \\
  \hline
ELS & Rosette Text Analytics & \textbf{31.7\%} & 38.5\% & 24.9\% & 40.4\% & 31.3\% \\
& Google Cloud NL & \textbf{43.2\%} & 44.9\% & 40.8\% & 45.5\% & 44.8\% \\ 
& TensiStrength & \textbf{44.2\%} & 52.1\% & 40.8\% & 43.6\% & 44.1\%  \\

 & mTurk crowdworkers & \textbf{74.7\%}  & 77.9\% & 71.7\% & 67.5\% & 80.5\% \\
  \\
\end{tabular}
}
\end{center}
\end{table*}

\citet{SamekiGeMaGuBe16} proposed a decision tree approach for computing the number of crowdworkers who should analyze a tweet based on the difficulty of the task. 
Tweets are labeled by 2, 3, 5, or 7 workers based on the difficulty of the task and the level of disagreement between the crowdworkers.
The model computes the number of workers based on how long a tweet is, the presence of a link in a tweet, and the number of present sarcasm signals.
Sarcasm is often used in political tweets and causes disagreement between the crowdworkers. The tweets that are deemed to be
sarcastic by the decision tree model, 
are expected to be more difficult to annotate, and hence are allocated more crowdworkers to work on. 


We conducted two sets of experiments. In the first set, we used \citet{tensi}, \citet{gcnlp}, and \citet{rosette}, for entity-level sentiment analysis; in the second set, \citet{gcnlp}, \citet{spaCy}, \citet{StanfordNER}, \citet{CogCompNLP}, and \citet{twitternlp11}, \citet{rosette} for named-entity recognition.


In the experiments that we conducted with TwitterNLP for named-entity recognition, we worked with the default values of the model. Furthermore, we selected the 3-class  Stanford NER  model, which uses the classes ``person,'' ``organization,'' and ``location'' because it resulted in higher accuracy compared to the 7-class model. For CogComp-NLP NER we used Ontonotes 5.0 NER model~\citep{WeischedelPaMaHo13}. For spaCy NER we used the `en\_core\_web\_lg' model.

We report the experimental results for our two tasks in terms of the correct classification rate (CCR).  For sentiment analysis, we have a three-class problem (positive, negative, and neutral), where the classes are mutually exclusive. The CCR, averaged for a set of tweets, is defined to be the number of correctly-predicted sentiments over the number of groundtruth sentiments in these tweets. 
For NER, we consider that each tweet may reference up to four candidates, i.e., targeted entities. The CCR, averaged for a set of tweets, is the number of correctly predicted entities (candidates) over the number of groundtruth entities (candidates) in this set.

\section{Results and Discussion}
\label{sec:results}

The dataset of 1,000 randomly selected tweets contains more than twice as many tweets about Trump than about the other candidates.  In the named-entity recognition experiment, the average CCR of crowdworkers was 98.6\%, while the CCR of the automated systems ranged from 77.2\% to 96.7\%. For four of the  automated systems, detecting the entity Trump was more difficult than the other entities (e.g., spaCy 72.7\% for the entity Trump vs.\ above 91\% for the other entities).  
An example of incorrect NER is shown in Figure~\ref{fig1} top.
The difficulties the automated tools had in NER may be explained by the fact that the tools were not trained on tweets, except for TwitterNLP, which was not in active development when the data was created~\citep{FarzindarIn15}.

In the sentiment analysis experiments, we found that a tweet may contain multiple sentiments. The groundtruth labels contain 210 positive sentiments, 521 neutral sentiments, and 305 negative sentiments to the candidates.  
We measured the CCR, across all tweets, to be 31.7\% for Rosette Text Analytics, 43.2\% for Google Cloud, 44.2\% for TensiStrength, and 74.7\% for the crowdworkers.  This means the difference between the performance of the tools and the crowdworkers is significant -- more than 30 percent points.

Crowdworkers correctly identified 62\% of the neutral, 85\% of the positive, and 92\% of the negative sentiments. Google Cloud correctly identified 88\% of the neutral sentiments, but only 3\% of the positive, and 19\% of the negative sentiments.  TensiStrength correctly identified 87.2\% of the neutral sentiments, but 10.5\% of the positive, and 8.1\% of the negative sentiments. Rosette Text Analytics correctly identified 22.7\% of neutral sentiments, 38.1\% of negative sentiments and 40.9\% of positive sentiments. The lowest and highest CCR pertains to tweets about Trump and Sanders for both Google Cloud and TensiStrength, Trump and Clinton for Rosette Text Analytics, and Clinton and Cruz for crowdworkers. An example of incorrect ELS analysis is shown in Figure~\ref{fig1} bottom.

\section{Conclusions and Future Work}

Our results show that existing NLP systems cannot accurately perform sentiment analysis of political tweets in the dataset we experimented with.  Labeling by humans, even non-expert crowdworkers, yields accuracy results that are well above the results of existing automated NLP systems.  In future work we will therefore use a crowdworker-labeled dataset to train a new machine-learning based NLP system for tweet analysis.   We will ensure that the training data is balanced among classes.  Our plan is to use state-of-the-art deep neural networks   
and compare their performance for entity-level sentiment analysis of political tweets. 


\clearpage
\section*{Acknowledgments}
Partial support of this work by the Hariri Institute for Computing and Computational Science \& Engineering at Boston University (to L.G.) and a Google Faculty Research Award (to M.B. and L.G.) is gratefully acknowledged. Additionally, we would like to thank Daniel Khashabi for his help in running the CogComp-NLP Python API and Mike Thelwal for his help with TensiStrength. We are also grateful to the Stanford NLP group for clarifying some of the questions we had with regards to the Stanford NER tool. 



\bibliographystyle{acl_natbib}
\bibliography{naaclhlt2018}

\appendix

\end{document}